\documentclass{article}
\usepackage{ijcai24}

\usepackage{times}
\usepackage{soul}
\usepackage{url}
\usepackage[hidelinks]{hyperref}
\usepackage[utf8]{inputenc}
\usepackage[small]{caption}
\usepackage{graphicx}
\usepackage{amsmath}
\usepackage{amsthm}
\usepackage{booktabs}
\usepackage{algorithm}
\usepackage{algorithmic}
\usepackage[switch]{lineno}

\usepackage{arydshln}
\usepackage{longtable}
\usepackage{xcolor}
\usepackage{amssymb}
\usepackage{adjustbox}
\usepackage{subfigure}
\usepackage{blindtext}
\usepackage{enumerate}
\usepackage{bm}
\usepackage{dsfont}
\usepackage{array}
\usepackage{multirow}
\usepackage{multicol}
\usepackage{hhline}
\usepackage{colortbl}

\newcolumntype{L}[1]{>{\raggedright\arraybackslash}p{#1}}
\newcolumntype{C}[1]{>{\centering\arraybackslash}p{#1}}
\newcolumntype{R}[1]{>{\raggedleft\arraybackslash}p{#1}}

\title{More is Better: Deep Domain Adaptation with Multiple Sources}

\author{
Sicheng Zhao$^1$\and
Hui Chen$^1$\and
Hu Huang$^3$\thanks{Corresponding Author.}\and
Pengfei Xu$^4$\and
Guiguang Ding$^{1,2}$\\
\affiliations
$^1$BNRist, Tsinghua University, China\ $^2$School of Software, Tsinghua University, China\\ $^3$Peking University Shenzhen Graduate School, China\ $^4$Didi Chuxing, China\\
\emails
\{schzhao, huichen, dinggg\}@tsinghua.edu.cn, h.huang@pku.edu.cn, xupengfeipf@didiglobal.com
}

\begin{document}

\maketitle

\begin{abstract}
In many practical applications, it is often difficult and expensive to obtain large-scale labeled data to train state-of-the-art deep neural networks. Therefore, transferring the learned knowledge from a separate, labeled source domain to an unlabeled or sparsely labeled target domain becomes an appealing alternative. However, direct transfer often results in significant performance decay due to \emph{domain shift}. Domain adaptation (DA) aims to address this problem by aligning the distributions between the source and target domains. Multi-source domain adaptation (MDA) is a powerful and practical extension in which the labeled data may be collected from multiple sources with different distributions. In this survey, we first define various MDA strategies. Then we systematically summarize and compare modern MDA methods in the deep learning era from different perspectives, followed by commonly used datasets and a brief benchmark. Finally, we discuss future research directions for MDA that are worth investigating. 
\end{abstract}

\section{Background and Motivation}
\label{sec:Motivation}

The availability of large-scale labeled training data, such as ImageNet, has enabled deep neural networks (DNNs) to achieve remarkable success in many learning tasks, ranging from computer vision to natural language processing~\cite{wilson2020survey,zhao2022review}. However, in many practical applications, obtaining sufficient labeled data is often expensive, time-consuming, and even impossible. For example, only experts can provide reliable annotations for fine-grained categories; annotating one Cityscapes image on the pixel level takes about 90 minutes~\cite{zhao2022review}.

\begin{figure}[!t]
\begin{center}
\centering \includegraphics[width=0.98\linewidth]{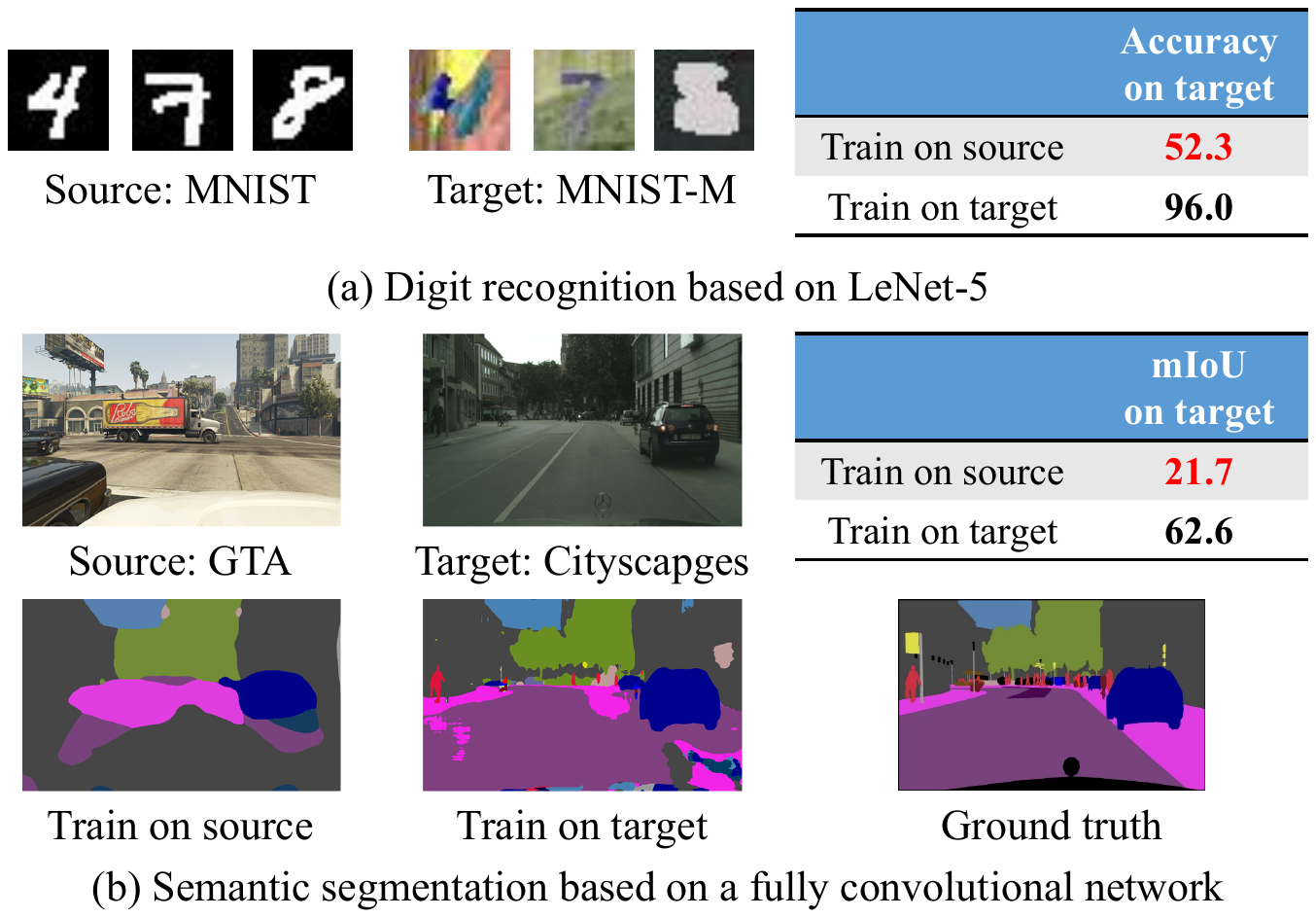}
\caption{Examples of \emph{domain shift} in the single-source scenario. The models trained on the labeled source domain do not perform well when directly transferring to the target domain.}
\label{fig:SingleDomainShift}
\end{center}
\end{figure}

One potential solution is to transfer the model trained on a separate, labeled source domain to the desired unlabeled or sparsely labeled target domain. But as Figure~\ref{fig:SingleDomainShift} illustrates, the \textbf{direct transfer of models across domains might lead to poor performance}, which is often termed negative transfer~\cite{zhang2023survey}. Figure~\ref{fig:SingleDomainShift}(a) shows that even for the simple task of digit recognition, training a LeNet-5 model on the MNIST source and directly transferring to the MNIST-M target leads to a classification accuracy decrease from 96.0\% to 52.3\%. Figure~\ref{fig:SingleDomainShift}(b) shows a more realistic example of training a semantic segmentation model on a synthetic source dataset GTA and conducting pixel-wise segmentation on a real target dataset Cityscapes. The poor performance of direct transfer across domains stems from \emph{domain shift}~\cite{zhao2022review}: the joint probability distributions of observed data and labels are different in the two domains.

\begin{figure}[!t]
\begin{center}
\centering \includegraphics[width=0.98\linewidth]{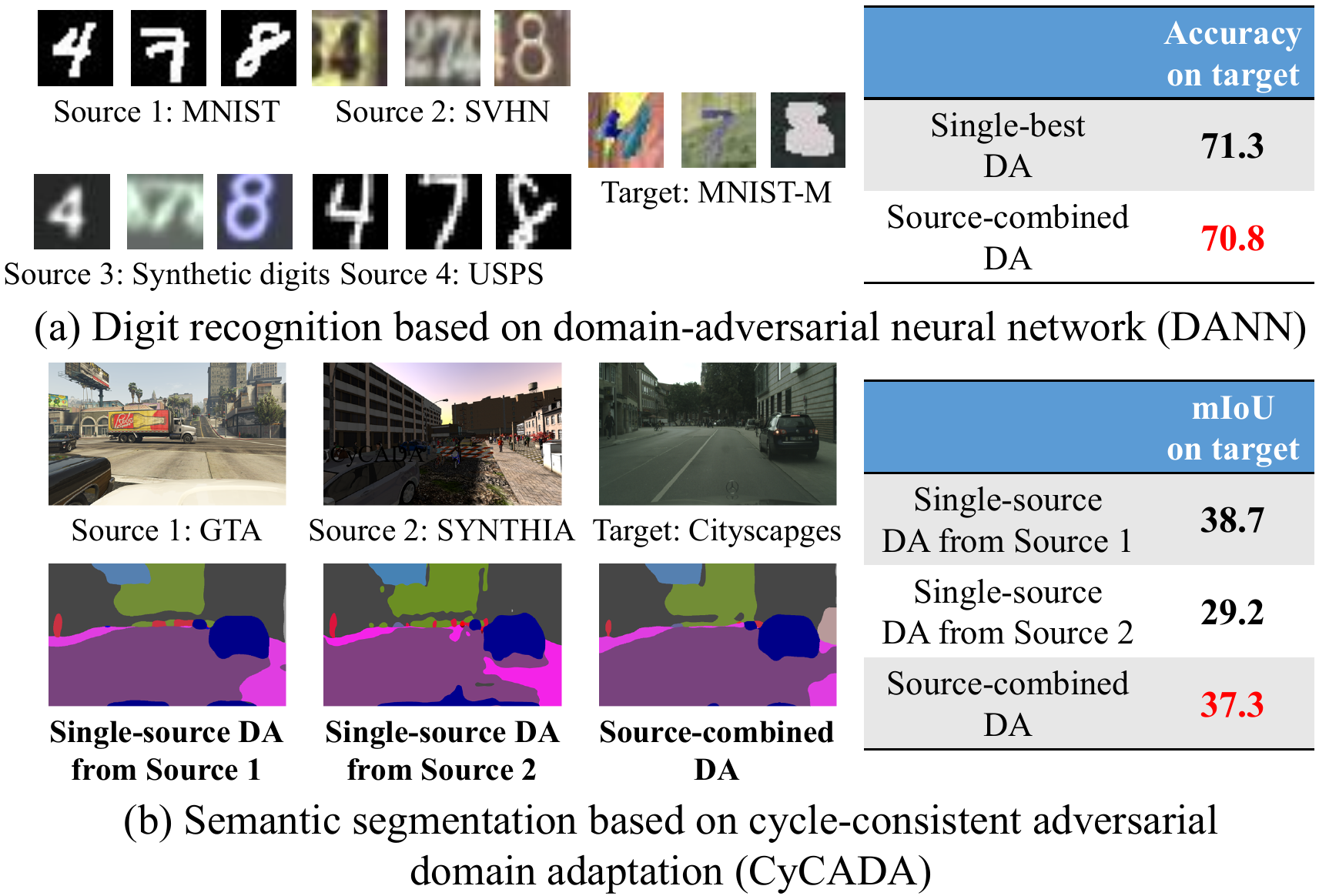}
\caption{Examples of \emph{domain shift} in the multi-source scenario. Simply combining multiple source domains into one source and directly performing SDA does not guarantee better performance compared to just using the best individual source.}
\label{fig:MultiDomainShift}
\end{center}
\end{figure}

The phenomenon of domain shift motivates the research on domain adaptation (DA), which aims to learn a model from a labeled source domain that can generalize well to a different but related target domain. Existing DA methods mainly focus on the single-source scenario. In the deep learning era, recent single-source DA (SDA) methods usually employ a conjoined architecture with two branches to respectively represent the models for the source and target domains. One branch aims to learn a task model based on the labeled source data using corresponding task losses, such as cross-entropy loss for classification. The other branch aims to deal with the domain shift by aligning the target and source domains. Based on the alignment strategies, deep SDA methods can be classified into different categories. The two most popular categories are 1) learning domain-invariant feature representations through discrepancy-based methods, adversarial discriminative methods, or self-supervision-based methods and 2) generating intermediate domains through generative models, such as Generative Adversarial Network (GAN). Other categories include ensemble methods, target discriminative methods, and so on. Please refer to~\cite{wilson2020survey,zhao2022review} for more details.

In practice, \textbf{the labeled data may be collected from multiple sources with different distributions}~\cite{sun2015survey}. 
In such cases, the aforementioned SDA methods could be trivially applied by combining multiple sources into a single one: an approach we refer to as \emph{source-combined DA}. However, source-combined DA might result in poorer performance than simply using one of the sources. As illustrated in Figure~\ref{fig:MultiDomainShift}, the accuracy on the best single-source digit recognition adaptation using DANN~\cite{ganin2016domain} is 71.3\%, while the source-combined accuracy drops to 70.8\%. Because the domain shift not only exists between each source and the target but also exists among different sources, the source-combined data may interfere with each other during the learning process~\cite{zhao2021madan}. Therefore, multi-source domain adaptation (MDA) is required to effectively leverage all the available data from different sources.

The early MDA methods mainly focus on shallow models~\cite{sun2015survey},
either learning a latent feature space for different domains or combining pre-learned source classifiers. Recently, the emphasis on MDA has shifted to deep learning architectures. In this paper, we systematically survey recent progress on deep learning-based MDA, summarize and compare the correlations and differences among these approaches, and discuss potential future directions.

\section{Problem Definition}
\label{sec:Definition}
In the typical MDA setting, there are multiple source domains $S_1,S_2,\cdots,S_M$ ($M$ is the number of sources) and one target domain $T$. Suppose the observed data and corresponding labels in the $i^{\text{th}}$ source $S_i$ drawn from distribution $p_i(\mathbf{x}, \mathbf{y})$ are $\mathbf{X}_i=\{\mathbf{x}_i^j\}_{j=1}^{N_i}$ and $\mathbf{Y}_i=\{\mathbf{y}_i^j\}_{j=1}^{N_i}$, respectively, where $N_i$ is the number of source samples. Let $\mathbf{X}_T=\{\mathbf{x}_T^j\}_{j=1}^{N_T}$ and $\mathbf{Y}_T=\{\mathbf{y}_T^j\}_{j=1}^{N_T}$ denote the target data and corresponding labels drawn from the target distribution $P_T(\mathbf{x},\mathbf{y})$, where $N_T$ is the number of target samples.

Suppose the number of labeled target samples is $N_{TL}$, the MDA problem can be classified into different categories:
\begin{itemize}
\item \emph{unsupervised MDA}, when $N_{TL}=0$;
\item \emph{fully supervised MDA}, when $N_{TL}=N_T$;
\item \emph{semi-supervised MDA}, otherwise.
\end{itemize}

Suppose $\mathbf{x}_i^j\in \mathds{R}^{d_i}$ and $\mathbf{x}_T^j\in \mathds{R}^{d_T}$ are an observation in $S_i$ and $T$, respectively, we can classify MDA into:
\begin{itemize}
\item \emph{homogeneous MDA}, when $d_1=\cdots=d_M=d_T$;
\item \emph{heterogeneous MDA}, otherwise.
\end{itemize}

Suppose $\mathcal{C}_i$ and $\mathcal{C}_T$ are the label set for source $S_i$ and target $T$, we can define different MDA strategies:
\begin{itemize}
\item \emph{closed-set MDA}, when $\mathcal{C}_1=\cdots=\mathcal{C}_M=\mathcal{C}_T$;
\item \emph{open-set MDA}, for all $\mathcal{C}_i$, $\mathcal{C}_i \cap \mathcal{C}_T \subset \mathcal{C}_T$;
\item \emph{partial MDA}, when $\mathcal{C}_T \subset (\mathcal{C}_1 \cup \mathcal{C}_2 \cup \cdots \cup \mathcal{C}_M)$;
\item \emph{union-set MDA}, $\mathcal{C}_1 \cup \mathcal{C}_2 \cup \cdots \cup \mathcal{C}_M = \mathcal{C}_T$;
\item \emph{universal MDA}, without prior knowledge on label sets;
\end{itemize}
where $\cap$, $\cup$, and $\subset$ respectively indicate the intersection set, union set, and proper subset between two sets.

Suppose the number of labeled source samples is $N_{iL}$ for source $S_i$, the MDA problem can be classified into:
\begin{itemize}
\item \emph{strongly supervised MDA}, when $N_{iL}=N_i$ for all $i$;
\item \emph{few-shot MDA}, when $N_{iL}<<N_i$ for all $i$.
\end{itemize}

Based on the data privacy and security constraints, the MDA task can be classified into:
\begin{itemize}
\item \emph{centralized MDA}, when $\mathbf{X}_i (i=1,\cdots,M)$ and $\mathbf{X}_T$ are available and shared during adaptation;
\item \emph{federated MDA}, when $\mathbf{X}_i (i=1,\cdots,M)$ and $\mathbf{X}_T$ are available but unshared during adaptation;
\item \emph{source-free MDA}, the trained source models are available but the source data cannot be accessed.
\end{itemize}

Multi-target MDA refers to adapting to multiple target domains simultaneously. When the target data is unavailable during training, the task becomes domain generalization.



\section{Deep MDA Taxonomy}
\label{sec:MDA}

Existing methods on deep MDA primarily focus on the unsupervised, homogeneous, closed-set, strongly supervised, centralized, one target, and target data available settings. That is, there is one target domain, the target data is unlabeled but available during adaptation, the source data is fully labeled and available during adaptation, the source and target data are observed in the same data space, and the label sets of all sources and the target are the same. In this paper, we will focus on MDA methods under these settings and will also briefly introduce other specific MDA settings.

There is some theoretical analysis to support existing MDA algorithms based on different statistical learning frameworks, such as VC dimension, Rademacher complexity, information theory, etc.
Most theories are based on the seminal theoretical model~\cite{blitzer2008learning,ben2010theory}, with an upper bound derived from VC dimension. \citeauthor{mansour2009mda}~\shortcite{mansour2009mda} assumed that the target distribution can be approximated by a convex combination of the $M$ source distributions and provided a distribution-weighted hypothesis combination rule. Moreover, tighter cross-domain generalization bounds and more accurate measurements of domain discrepancy can provide intuitions to derive effective MDA algorithms. 
\citeauthor{hoffman2018algorithms}~\shortcite{hoffman2018algorithms} derived a novel bound using DC-decomposition and calculated more accurate combination weights of the non-convex optimization situation. \citeauthor{zhao2018adversarial}~\shortcite{zhao2018adversarial} extended the generalization bound of the seminal theoretical model to multiple sources under both classification and regression settings. Besides the domain discrepancy between the target and each source~\cite{hoffman2018algorithms,zhao2018adversarial}, \citeauthor{li2018extracting}~\shortcite{li2018extracting} also considered the relationship between pairwise sources and derived a tighter bound on the weighted multi-source discrepancy. Based on this bound, more relevant source domains can be picked out. 
A finite-sample error bound is derived~\cite{wen2020domain} using sample-based Rademacher complexity, a generally tighter complexity measure than VC dimension. The attention mechanism is proven to automatically reduce the negative effects caused by unrelated domains~\cite{zuo2021attention}. 

The label distribution has also been theoretically studied. To deal with the inconsistency between the source and target label distributions, \citeauthor{redko2019optimal}~\shortcite{redko2019optimal} provided a generalization bound when transferring under target shift. Later, the similarity of the semantic conditional distribution is designed to aggregate multiple sources~\cite{shui2021aggregating}. Mutual information in the information theory provides another theoretical solution. \citeauthor{park2021information}~\shortcite{park2021information} proved that the mutual information between latent representations and domain labels is closely related to the explicit optimization of the divergence between the source and target domains. Recently, \citeauthor{chen2023algorithm}~\shortcite{chen2023algorithm} provided algorithm-dependent generalization bounds characterized by the mutual information between the parameters and the data. 

\begin{figure}[!t]
\begin{center}
\centering \includegraphics[width=0.98\linewidth]{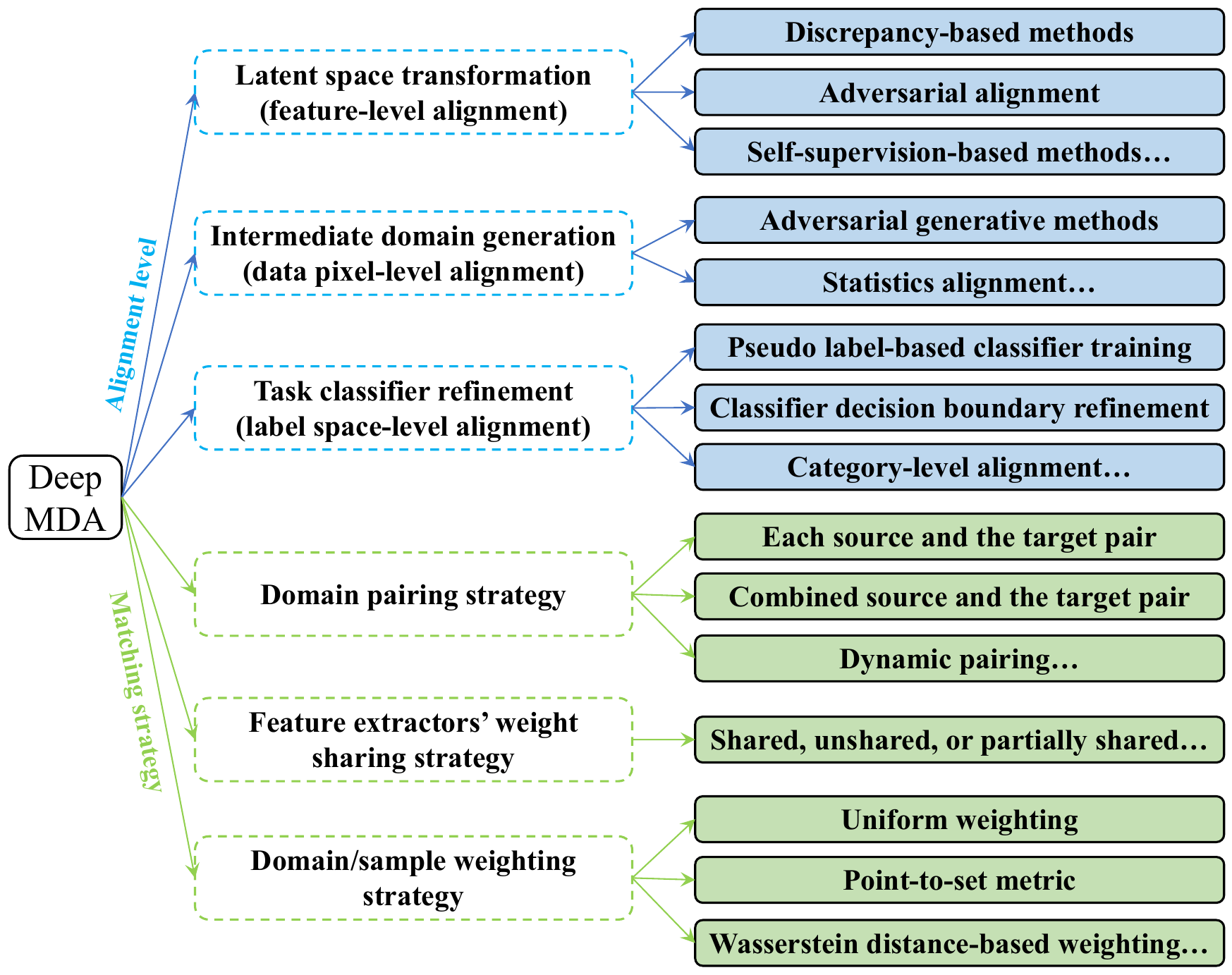}
\caption{Taxonomy of deep MDA methods.}
\label{fig:Taxonomy}
\end{center}
\end{figure}

\begin{figure*}[!t]
\begin{center}
\centering \includegraphics[width=0.98\linewidth]{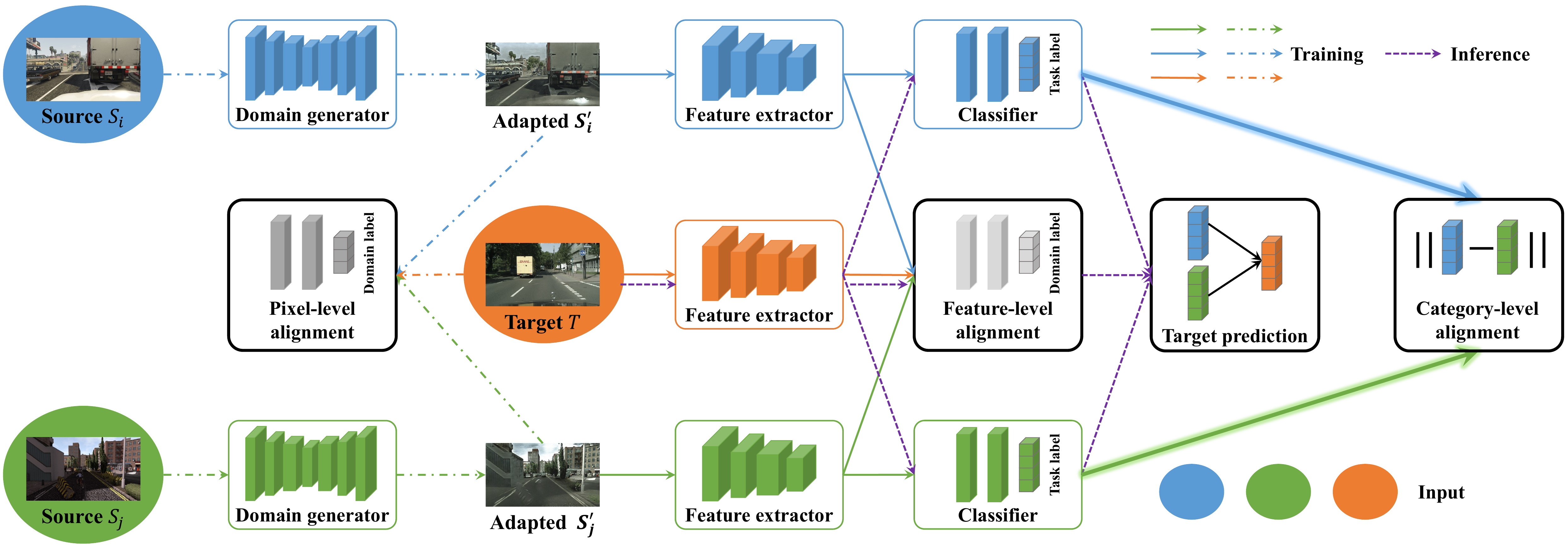}
\caption{A widely employed MDA framework. The solid arrows, dashed dot arrows, and bold arrows indicate the training of latent space transformation, intermediate domain generation, and task classifier refinement, respectively. The dashed arrows indicate the reference process.}
\label{fig:Framework}
\end{center}
\end{figure*}

Typically, some task models (\textit{e.g.}, classifiers) are learned based on the labeled source data with corresponding task loss. Meanwhile, specific alignments among the source and target domains are conducted to bridge the domain shift so that the learned task classifiers can be better generalized to the target domain. Based on the alignment methodologies on different levels, we can classify MDA into different categories, as shown in Figure~\ref{fig:Taxonomy}. \emph{Latent space transformation} that performs feature-level alignment tries to learn domain-invariant features of different domains. \emph{Intermediate domain generation} aims to align different domains on the data pixel level by explicitly generating an intermediate adapted domain for each source that is indistinguishable from the target domain. \emph{Task classifier refinement} considers the label distribution variance so that the learned task classifiers can better fit the target domain. Latent space transformation and intermediate domain generation focus on bridging the covariate shift, \textit{i.e.}, the conditional distributions of labels given an observation are the same across domains, but the marginal distributions of observations are different. Task classifier refinement aims to address the label shift where the marginal label distributions differ across domains. Figure~\ref{fig:Framework} illustrates a widely employed framework of existing MDA methods. Meanwhile, several matching strategies should be considered to deal with the case of multiple sources and one target, including domain pairing strategy, feature extractors' weight sharing strategy, and domain/sample weighting strategy.

\section{MDA Alignment Methodologies}
\label{sec:MDA_Alignment}
\subsection{Latent Space Transformation}
\label{ssec:SpaceTransformation}

The common methods for aligning the latent spaces of different domains, \textit{i.e.}, learning domain-invariant feature representations, include discrepancy-based methods, adversarial alignment, and self-supervision-based methods.

\textbf{Discrepancy-based methods} measure the discrepancy of the latent spaces (typically features) from different domains by optimizing some specific discrepancy losses, such as maximum mean discrepancy (MMD)~\cite{guo2018multi,zhu2019aligning,li2022multi,wang2022self}, R\'enyi-divergence~\cite{hoffman2018algorithms}, moment distance~\cite{peng2019moment,fu2021partial}, weighted moment distance~\cite{zuo2021attention}, correlation alignment~\cite{karisani2022multiple}, relation alignment loss~\cite{wang2020learning}, contrastive domain discrepancy~\cite{kang2022contrastive}, $\mathcal{L}_1$ distance~\cite{cai2023exploring}, and Wasserstein distance~\cite{chen2023algorithm}. \citeauthor{guo2020multi}~\shortcite{guo2020multi} claimed that different discrepancies or distances can only provide specific estimates
of domain similarities and thus they considered the mixture of multiple distances. Minimizing the discrepancy to align the features among the source and target domains usually does not introduce any new to-be-learned parameters. Instead of using the above-mentioned explicit discrepancy metrics, \citeauthor{li2018adaptive}~\shortcite{li2018adaptive} proposed Adaptive Batch Normalization (AdaBN) to implicitly minimize the discrepancy between the source and target domains by aligning the BN statistics, such as the moving average mean and variance of all BN layers.


\textbf{Adversarial alignment} employs a discriminator to align the features by making them indistinguishable. Some representative optimized objectives include GAN loss~\cite{xu2018deep,zhao2021madan,nguyen2021stem,liu2022two,zhang2022multi}, weighted GAN loss~\cite{yang2020curriculum,yin2022universal}, attention-guided GAN loss~\cite{gong2021mdalu}, $\mathcal{H}$-divergence~\cite{zhao2018adversarial,park2021information}, $d_\mathcal{A}$ distance~\cite{wen2020domain}, Wasserstein distance~\cite{li2018extracting,zhao2020multi}, least squares loss and focal loss~\cite{yao2021multi},  weighted conditional Wasserstein distance~\cite{shui2021aggregating}, cross-entropy loss~\cite{wu2022adversarial,zhou2022prototype,belal2024multi},  $\mathcal{L}_2$ 0loss~\cite{yao2021multisource}, and metric constraint loss~\cite{ren2022multi}. These methods aim to confuse the discriminator's ability to distinguish whether the features from multiple sources were drawn from the same distribution. Compared with GAN loss and $\mathcal{H}$-divergence, Wasserstein distance can provide
more stable gradients even when the target and source distributions do not overlap~\cite{zhao2020multi}. Instead of directly taking the output of feature extractors as input into the discriminator, \citeauthor{zhao2021curriculum}~\shortcite{zhao2021curriculum} employed another generator to generate adapted representations, which are input into the discriminator for alignment. The discriminator is often implemented as a network, which leads to new parameters that need to be learned.

\textbf{Self-supervision-based methods} typically combine some auxiliary self-supervised learning (pretext) tasks with the target task. Some commonly used pretext tasks include reconstruction, image rotation, and patch location prediction~\cite{zhao2022review}. In~\cite{yuan2022self}, the domain information is considered as the labels of the pretext task. That is, domain classification plays the role of pretext task. 

\subsection{Intermediate Domain Generation}
\label{ssec:DomainGeneration}
Feature-level alignment only aligns high-level information, which is insufficient for fine-grained predictions, such as pixel-wise 
semantic segmentation~\cite{zhao2019multi,zhao2021madan}. Generating an intermediate adapted domain with pixel-level alignment, typically via GAN, diffusion models, and their variants, can help address this problem. \citeauthor{zhao2021madan}~\shortcite{zhao2021madan} used CycleGAN to generate an adapted domain for each source and proposed to aggregate different adapted domains using a sub-domain aggregation discriminator and a cross-domain cycle discriminator, where the pixel-level alignment is then conducted between the aggregated and target domains. Instead of taking the original source data as input to the generator, \citeauthor{lin2020multi}~\shortcite{lin2020multi} used a variational autoencoder to map all source and target domains to a latent space and then generated an adapted domain from the latent space. \citeauthor{zhao2021madan}~\shortcite{zhao2021madan} and \citeauthor{lin2020multi}~\shortcite{lin2020multi} showed that the semantics might change in the intermediate domain and that enforcing a semantic consistency before and after generation can help preserve the original semantics. \citeauthor{he2021multi}~\shortcite{he2021multi} proposed to translate the style of images in source domains to the style of the target domain by aligning the statistics distribution of pixel values in LAB color space. Besides the standard GAN loss, \citeauthor{kang2023structure}~\shortcite{kang2023structure} further enforced the minimization of a mutual information loss and a texture co-occurrence
loss to respectively constrain structural changes and encourage high-quality texture translation in local areas.

\subsection{Task Classifier Refinement}
\label{ssec:ClassifierRefinement}
Even after the source and target domains are aligned on the feature level and pixel level by latent space transformation and intermediate domain generation, the learned task classifiers on the labeled sources still cannot be guaranteed to perform well on the target. For example, the classifiers trained on different sources may result in misaligned predictions for the target samples that are close to the domain boundary; the class distributions are imbalanced across domains. Task classifier refinement aims to address these issues.

One intuitive method is \textbf{pseudo label-based classifier training} which employs the target pseudo labels for classifier training~\cite{venkat2020your,he2021multi}. A logit-based metric is designed to select reliable pseudo labels~\cite{venkat2020your}. 
\textbf{Classifier decision boundary refinement} tries to refine the decision boundary of the classifier to better fit the target domain. 
Specific classifier discrepancies, such as $\mathcal{L}_1$ loss~\cite{zhu2019aligning,peng2019moment,yao2021multi}, and divergence loss~\cite{nguyen2021stem,karisani2022multiple}, are minimized to encourage the predictions of the same target sample by different classifiers to be the same. Both intra-domain and inter-domain consistency losses based on $\mathcal{L}_1$ are considered~\cite{luo2022domain}. Boundary-sensitive alignment is proposed to assign different weights to the source-target alignments based on the cross-entropy loss of the labeled source samples~\cite{kang2022contrastive}. The dynamic classifier alignment method aligns classifiers derived from multi-view features through an automatic, sample-wise approach~\cite{li2023dynamic}.

\textbf{Category-level alignment} aligns the source and target domains, usually based on the feature representations, for each category. MADAN+ employs one discriminator for each category to align the features of different domains in an adversarial manner~\cite{zhao2021madan}. Differently, category-level discrepancy-based alignment incorporates the class distribution prior information when computing the MMD discrepancy between different domains~\cite{wang2022self,li2022multi}. Since the target domain is unlabeled, obtaining reliable pseudo target labels plays an important role in category-level alignment.  Categorical prototypes are aligned across different domains based on minimax entropy loss~\cite{liu2022two}, reconstruction~\cite{zhou2022prototype}, and cosine similarity~\cite{belal2024multi}. The SIG model integrates class-aware conditional alignment in domain-invariant and label-relevant latent variables to address changes in label distributions across different domains~\cite{li2023subspace}.

\section{MDA Matching Strategies}
\subsection{Domain Pairing Strategy}
\label{ssec:DomainPairing}

There are different ways to group the target and multiple sources so that alignment can be performed. The most common method is to pairwise align the target with each source, such as~\cite{xu2018deep,zhao2018adversarial,zhao2020multi,karisani2022multiple}, to name a few. Since domain shift also exists among different sources, several methods also enforce pairwise alignment between all sources~\cite{li2018extracting,peng2019moment,yin2022universal}. Different adapted domains are first aggregated into one domain, which is then aligned with the target~\cite{zhao2021madan,li2021dynamic,venkat2020your,chen2023algorithm}. A multi-armed bandit controller is employed to dynamically select one source to learn an optimal trajectory and mixture of domains for adaptation~\cite{guo2020multi}. All the source and target domains are directly discriminated by a multi-class discriminator~\cite{park2021information}. A tensor-low-rank constraint is imposed on the prototypical similarity matrices to explore the high-order relationships among different domains~\cite{li2021t}. 

All the above methods assume that the domain label information is available. However, this might not hold in practice. In such cases, a naive approach is to directly combine all the sources into one to align with the target~\cite{yang2020curriculum,zhao2021curriculum,li2021dynamic}. 

\subsection{Feature Extractors' Weight Sharing Strategy}
\label{ssec:WeightSharing}
Most MDA methods employ shared feature extractors to learn domain-invariant features, such as~\cite{zhao2018adversarial,peng2019moment,kang2022contrastive,wang2022self,li2023subspace,belal2024multi}. However, domain invariance may be detrimental to the discriminative power and it might be difficult to extract the same domain-invariant features for all domains. Besides the same feature extractor, $M$ domain-specific feature extractors are trained so that each source and the target can be better aligned~\cite{zhu2019aligning}, while only $M$ domain-specific feature extractors are considered in~\cite{he2021multi}. One feature extractor is first pre-trained for each source and then the target is mapped into the feature space of each source~\cite{zhao2020multi}. The source domains share a feature extractor and the target uses a separate one~\cite{wu2022target}. The source domains with similar characteristics based on coordination share the same feature extractor~\cite{karisani2022multiple}. Low-level features that are important for localization share the same extractor across domains with strong alignment, while high-level features that are important for object recognition use specific extractors with weak alignment~\cite{yao2021multi,zhang2022multi}. Although unshared feature extractors can better align the target and sources in the latent space, this substantially increases the number of parameters in the model.


\subsection{Domain/Sample Weighting Strategy}
\label{ssec:DomainWeighting}
Some early deep MDA methods treat different source domains and source samples equally. For example, only one task classifier is trained for all the source domains even with available domain label information~\cite{zhao2018adversarial}. However, theis ignores the different correlations between multiple sources and the target as well as the different similarities between source and target samples. Assigning different source domains and source samples with optimal weights would result in better alignment and generalization. One common domain weighting strategy is to combine the target predictions by multiple source classifiers with different weights during inference, such as uniform weight~\cite{zhu2019aligning}, perplexity score based on adversarial loss~\cite{xu2018deep}, point-to-set (PoS) metric using Mahalanobis distance~\cite{guo2018multi}, relative error based on source-only accuracy~\cite{peng2019moment},  Wasserstein distance based weights~\cite{zhao2020multi}, and majority voting~\cite{karisani2022multiple}.


The other domain weighting strategy aims to adjust the importance weights of source domains during training~\cite{wen2020domain}. An uncertainty-aware weighting strategy is designed to adaptively assign weights to different source domains and samples~\cite{li2021t}. Based on the attention- or discriminator-based domain correlations that measure the probabilities of target samples belonging to each source domain, weighted moment distance is designed to pay more attention to the source domains with higher similarities during alignment~\cite{zuo2021attention,yao2021multi,li2023dynamic}. Domain weights are derived based on the upper bound or the gradient norm bound~\cite{shui2021aggregating,chen2023algorithm}.

Sample weighting has also been explored recently. Based on a manually selected Wasserstein distance threshold, the source samples that are closer to the target samples are distilled to fine-tune the source classifiers~\cite{zhao2020multi}. During adversarial alignment, the source samples in a batch are assigned with different weights by model-based curriculum with a source selection network~\cite{yang2020curriculum,zhao2021curriculum} or by model-free curriculum based on the output of domain discriminators~\cite{zhao2021curriculum,liu2022two,zhang2022multi}. The source samples are re-weighted based on the source-target relevance obtained by the nearest neighbor algorithm~\cite{wu2022target}. A weighted loss term with a general self-paced regularizer is employed to optimize both model parameters and sample weights~\cite{wang2022self}. The source features are weighted by target similarity evaluated by the normalized domain discriminator output~\cite{liu2022two}.

\section{Other Specific MDA Settings}
\label{sec:OtherSettings}


\textbf{Heterogeneous MDA.} Transforming the heterogeneous input from different domains into a shared latent space with the same feature dimensions is an intuitive solution. The feature extractor for each domain contains two networks~\cite{yao2021multisource}. One is domain-specific with unshared structures and parameters. The other has identical structures with a $\mathcal{L}_1$-based parameter consistency constraint, which aims to make the parameters consistent across domains.

\textbf{Open-set, partial, and universal MDA.} The core of open-set MDA lies in how to align the known classes between the sources and the target while making the unknown classes separable. \citeauthor{rakshit2020multi}~\shortcite{rakshit2020multi} studied the case that all sources share the same label set, while the target contains unknown classes. A binary pseudo classifier with a margin loss is trained to classify whether a target sample belongs to the unknown class. In partial MDA, the relevant samples in the shared classes need to be emphasized while the irrelevant ones in the sources need to be ignored. In~\cite{fu2021partial}, a selection vector is first derived to refine the extracted features and then the selected features are aligned on both the domain level and the category level using $\mathcal{L}_2$ loss and local relation alignment Loss, respectively. An attention-guided adversarial alignment and pseudo-label supervision fusion mechanisms are designed to effectively harness both labeled and unlabeled data across domains~\cite{gong2021mdalu}. The main challenge in universal MDA is to identify the common label set among each source and the target and keep the model scalable as the number of source domains increases. A pseudo-margin vector that measures the confidence in assigning a target sample to the pseudo label helps to identify the common label set and to adversarially align the distributions of multiple sources and the unknown target~\cite{yin2022universal}.

\textbf{Federated and source-free MDA.} In the pursuit of privacy protection, federated and source-free MDA that both belongs to decentralized MDA has emerged. 
In federated MDA, the model parameters at each node are trained separately with convergence at different rates. To address these challenges, \citeauthor{peng2020federated}~\shortcite{peng2020federated} first introduced federated adversarial adaptation to optimize the $\mathcal{H}$-divergence without accessing data, while employing a dynamic attention mechanism and leveraging representation disentanglement to mitigate negative transfer to the target domain. To further reduce the communication costs and prevent privacy leakage attacks, \citeauthor{feng2021kd3a}~\shortcite{feng2021kd3a} proposed a multi-source knowledge distillation method and a dynamic weighting strategy to respectively obtain high-quality domain consensus and identify the irrelevant sources. 
Based on whether the parameters of pre-trained source models are available, source-free MDA (SFMDA) can be divided into white box and black box. A common white box SFMDA pipeline involves using pseudo labels, which often introduce noise that impacts the adaptation result. Several pseudo-label denoising strategies have been proposed, such as pseudo label selection based on self-supervised clustering~\cite{ahmed2021unsupervised} and confidence query~\cite{shen2023balancing}, and confident-anchor-induced pseudo label generation~\cite{dong2021confident}. A source importance estimation module~\cite{han2023discriminability} and a concurrent subsidiary supervision module~\cite{kundu2022concurrent} are designed to integrate into existing models to enhance the adaptation performance. For black box SFMDA, the knowledge from source predictors is first distilled to a customized target model, which is then fine-tuned with the target data~\cite{liang2022dine}.

%
%


\textbf{Meta-learning-based MDA.} Meta-learning aims to learn a general model from multiple tasks by induction. Incorporating meta-learning in MDA involves creating algorithms that can discern and leverage the underlying structure shared across different domains. \citeauthor{li2020online}~\cite{li2020online} proposed to use meta-learning to enhance the performance of existing MDA methods. Specifically, the multi-source domains are split into disjoint meta-training and meta-testing domains. The best initial condition that is suited to adapting between all source domains is searched for the sake of adapting well to the target domain.

\section{Applications, Datasets, and Benchmark}
\label{sec:Datasets}

MDA has been successfully applied in various areas, ranging from computer vision (CV) to natural language processing (NLP) and from speech recognition to industry diagnosis. Specifically, MDA has been studied in multiple learning tasks, such as image classification, object detection, and semantic segmentation in CV, as well as sentiment classification and machine translation in NLP. The datasets for evaluating MDA models usually contain multiple domains with different styles, such as \textit{synthetic} vs. \textit{real}, \textit{artistic} vs. \textit{sketchy}, which impose large domain shift. Here we briefly introduce some commonly used datasets in both CV and NLP areas.

Image classification is the most popular CV task for MDA. Some representative datasets include Digits-five for digit recognition, Office-31, Office-Caltech, ImageCLEF, PACS, Office-Home, and DomainNet for object classification. Existing MDA methods can obtain an accuracy that is higher than 90\% for some datasets (\textit{e.g.}, Digits-five, Office-31, Office-Caltech, and ImageCLEF) and even 100\% for some domains (\textit{e.g.}, Webcam and DSLR in Office-Caltech)~\cite{yuan2022self}. Office-Home and DomainNet~\cite{peng2019moment} are more challenging. Office-Home consists of about 15,500 images from 65 categories of everyday objects in office and home settings with 4 domains. DomainNet contains about 600K images from 345 categories with 6 domains: Clipart (\textit{Clp}), Infograph (\textit{Inf}), Painting (\textit{Pnt}), Quickdraw (\textit{Qdr}), Real (\textit{Rel}), and Sketch (\textit{Skt}). Sim2Real MDA is often performed for object detection and semantic segmentation. Commonly used synthetic domains include GTA and SYNTHIA, while Cityscapes, KITTI, and BDD100K are popular real domains.

\begin{table}[!t]
    \centering
    \resizebox{\linewidth}{!}{%
    \begin{tabular}{c|c c c c c c |c}
    \toprule
        ~ & Clp & Inf & Pnt & Qdr & Rel & Skt & Avg  \\ \hline
        Source-B & 39.6  & 8.2  & 33.9  & 11.8  & 41.6  & 23.1  & 26.4   \\ 
        Source-C & 47.6  & 13.0  & 38.1  & 13.3  & 51.9  & 33.7  & 32.9   \\ \hline
        DCTN & 48.6  & 23.5  & 48.8  & 7.2  & 53.5  & 47.3  & 38.2   \\ 
        MDAN & 52.4  & 21.3  & 46.9  & 8.6  & 54.9  & 46.5  & 38.4   \\ 
        M3SDA & 58.6  & 26.0  & 52.3  & 6.3  & 62.7  & 49.5  & 42.6   \\
        MDDA & 59.4  & 23.8  & 53.2  & 12.5  & 61.8  & 48.6  & 43.2   \\ 
        CMSS & 64.2  & 28.0  & 53.6  & 16.0  & 63.4  & 53.8  & 46.5   \\ 
        LtC-MSDA & 63.1  & 28.7  & 56.1  & 16.3  & 66.1  & 53.8  & 47.4   \\ 
        SImPAI & 66.4  & 26.5  & 56.6  & 18.9  & 68.0  & 55.5  & 48.6   \\ 
        T-SVDNet & 66.1  & 25.0  & 54.3  & 16.5  & 65.4  & 54.6  & 47.0   \\ 
        ABMSDA & 66.9  & 25.3  & 55.8  & 18.2  & 64.1  & 55.2  & 47.6   \\ 
        DEAL & 70.8  & 26.5  & 57.4  & 12.2  & 65.0  & 60.6  & 48.7   \\ 
        DRT+ST & 71.0 & 31.6 & 61.0 & 12.3 & 71.4 & \textbf{60.7}  & 51.3 \\
        PTMDA & 66.0  & 28.5  & 58.4  & 13.0  & 63.0  & 54.1  & 47.2   \\ 
        SPS & 70.8  & 24.6  & 55.2  & 19.4  & 67.5  & 57.6  & 49.2   \\ 
        SSG & 68.7  & 24.8  & 55.7  & 18.4  & 68.8  & 56.3  & 48.8   \\ 
        MSCAN & 69.3  & 28.0  & 58.6  & \textbf{30.3}  & \textbf{73.3}  & 59.5  & \textbf{53.2}   \\ 
        SIG	& \textbf{72.7} & \textbf{32.0}  & \textbf{61.5} & 20.5 & 72.4  & 59.5 & 53.0 \\
        \hline
        Oracle & 69.3  & 34.5  & 66.3  & 66.8  & 80.1  & 60.7  & 63.0  \\ \bottomrule
    \end{tabular}
    }
    \caption{Classification accuracy (\%) of different adaptation tasks on DomainNet based on ResNet-101. The best performance trained on the source domains is emphasized in bold.}
    \label{tab:DomainNet}
\end{table}

Sentiment classification is a widely-studied NLP task for MDA. Some representative datasets include Amazon Reviews, Media Reviews, Reviews-5, and Multilingual Amazon Reviews Corpus (MARC). As a relatively new dataset, MARC is a collection of Amazon reviews from four languages: German, English, French, and Japanese. For each language, there are three domains: Books, DVD, and Music. We can perform cross-domain adaptation for each language and cross-lingual adaptation for each domain.

To give readers a brief and quick understanding of how current MDA methods perform, we report the object classification accuracy of several adaptation tasks on DomainNet. One domain is selected as the target and the rest domains are used as multiple sources. The compared MDA methods include DCTN~\cite{xu2018deep}, MDAN~\cite{zhao2018adversarial}, M3SDA~\cite{peng2019moment}, MDDA~\cite{zhao2020multi}, CMSS~\cite{yang2020curriculum}, LtC-MSDA~\cite{wang2020learning}, SImPAI~\cite{venkat2020your}, T-SVDNet~\cite{li2021t}, ABMSDA~\cite{zuo2021attention}, DEAL~\cite{zhou2022prototype}, DRT+ST
~\cite{li2021dynamic}, PTMDA~\cite{ren2022multi}, SPS~\cite{wang2022self}, SSG~\cite{yuan2022self}, MSCAN~\cite{kang2022contrastive}, and SIG~\cite{li2023subspace}. For comparison, we report the results of source-only methods that are trained on the source domains and directly tested on the target domain under two settings: single-best (`Source-B') and source-combined (`Source-C'). The result of an oracle setting is also reported, where the classification model is both trained and tested on the target domain. The results of source-only and oracle methods can be viewed as a lower bound and an upper bound of MDA.

From the results in Table~\ref{tab:DomainNet}, we have the following observations: (1) Because of the domain gap, the source-only methods without any alignment perform the worst in most adaptation settings. (2) Through alignment on different levels, such as the feature level, the MDA methods can significantly improve the classification accuracy. As compared to Source-C, MSCAN achieves a 20.3\% performance gain. (3) For specific domains, such as Clp, some adaptation algorithms (\textit{e.g.}, SIG) can even outperform the oracle method, demonstrating the effectiveness and necessity of MDA. (4) On average, there is still a large performance gap between MDA methods and oracle, which calls for further research efforts to better reduce the domain gap.

\section{Conclusion and Future Directions}
\label{sec:Conclusion}

In this paper, we provided a systematic survey of recent MDA developments in the deep learning era. We motivated MDA, defined different strategies, and summarized the commonly used datasets with a brief benchmark. We classified existing methods into different categories and compared the representative ones technically and experimentally. We conclude this paper with several open research directions:

\textbf{Specific MDA strategy implementation.} There are many  MDA strategies defined from different aspects. In practice, we might encounter specific combinations of these strategies, such as union-set source-free MDA. Specific implementation of such an MDA strategy would likely yield better results than a one-size-fits-all MDA approach. 

\textbf{Multi-modal and time-series MDA.} The labeled source data may be of different modalities, such as LiDAR, radar, and image. Further research is needed to find techniques for fusing different data modalities in MDA. A further extension of this idea is to have varied modalities in different sources and the modalities contain time-series information.

\textbf{Incremental and test-time MDA.} Incremental MDA remains largely unexplored and may provide great benefits for real-world scenarios, such as updating deployed MDA models when new source or target data becomes available. Test-time MDA that allows more flexible inference time can provide inference unlabeled data for adaptation.

\textbf{Large-scale pre-training and self-learning for MDA.}
The rapid development of large-scale pre-training and self-learning has been witnessed recently, such as 
BERT and GPT. Employing such pre-training and self-learning techniques to explore their general representation for MDA is a promising direction.

\textbf{Theoretical and interpretable analysis.}
Theoretical and interpretable analysis can guide the design of MDA algorithms and provide novel insights to understand the pros and cons of MDA models. For example, we may derive a tighter upper bound, analyze whether every source domain and source sample contributes, and visualize the learned features to see if different categories are correctly classified.

\section*{Acknowledgements}
This work is supported by CCF-DiDi GAIA Collaborative Research Funds for Young Scholars and the National Natural Science Foundation of China (Nos. 61925107, 62021002).

\bibliographystyle{named}
\bibliography{cameraReady}

\end{document}